\documentclass[twoside,11pt]{article}
\usepackage{jair, theapa, rawfonts}

\usepackage{graphicx}
\usepackage{epsfig}
\usepackage{epstopdf}
\usepackage{subcaption}
\usepackage{array}

\usepackage{tikz}
\usetikzlibrary{positioning,shadows,arrows,snakes,shapes}

\ShortHeadings{An Overview of Probabilistic Graphical Models for Transfer Learning}
{XUAN, LU, \& ZHANG}
\firstpageno{1}

\begin{document}

\title{Bayesian Transfer Learning: An Overview of Probabilistic Graphical Models for Transfer Learning}

\author{\name Junyu Xuan \email Junyu.Xuan@uts.edu.au \\
       \name Jie Lu \email Jie.Lu@uts.edu.au\\
       \name Guangquan Zhang\email Guangquan.Zhang@uts.edu.au\\
       \addr Australia Artificial Intelligence Institute,
              Faculty of Engineering and Information Technology,
              University of Technology Sydney, PO Box 123, Broadway, NSW 2007, Sydney, Australia
 }


\maketitle

\begin{abstract}
Transfer learning where the behavior of extracting transferable knowledge from the source domain(s) and reusing this knowledge to target domain has become a research area of great interest in the field of artificial intelligence.
Probabilistic graphical models (PGMs) have been recognized as a powerful tool for modeling complex systems with many advantages, e.g., the ability to handle uncertainty and possessing good interpretability.
Considering the success of these two aforementioned research areas, it seems natural to apply PGMs to transfer learning.
However, although there are already some excellent PGMs specific to transfer learning in the literature, the potential of PGMs for this problem is still grossly underestimated.
This paper aims to boost the development of PGMs for transfer learning by 1)  examining the pilot studies on PGMs specific to transfer learning, i.e., analyzing and summarizing the existing mechanisms particularly designed for knowledge transfer; 2) discussing examples of real-world transfer problems where existing PGMs have been successfully applied; and 3) exploring several potential research directions on transfer learning using PGM.

\end{abstract}

\section{Introduction}
\label{Introduction}

Transfer learning is the behavior of extracting transferable knowledge from the source domain(s) and reusing this knowledge in the target domain, which is a natural human phenomenon, even for very young children \cite{Brown1988493}. A formal definition
is as follow \cite{5288526}:
``\emph{Given a source domain $\mathcal{D}_S=\{\mathcal{X}_S, P_S(X)\}$ and a target domain $\mathcal{D}_T=\{\mathcal{X}_T, P_T(X)\}$, transfer learning aims to improve the learning task in $\mathcal{D}_T$ by the help of $\mathcal{D}_S$, where $\mathcal{X}$ is feature space and $P(X)$ is the data distribution.}''
When $\mathcal{X}_S = \mathcal{X}_T$, it is homogeneous transfer learning; when $\mathcal{X}_S \neq \mathcal{X}_T$, it is heterogeneous transfer learning.
Note that transfer learning could be deemed as the aforementioned specified problem or the methodology to resolve this problem as well.
A classic motivating example is cross-domain (e.g., movie and computer domains) sentiment prediction for product reviews: 1) There are a large number of labeled product reviews in the movie domain so a classifier could be well-trained and applied to make predications in this domain; 2) The labeled reviews for a new computer are seldom sufficient enough to train a  classifier for further sentiment prediction; 3) A straightforward idea is to directly apply the classifier from the movie domain to the new computer domain considering the \emph{similarity} between two domains (e.g., people tend to use similar words to express their likes or dislikes across different products), but it does not always work well and may lead to \emph{negative transfer} \cite{weiss2016survey} because of their \emph{differences} in different contexts (e.g., \emph{touch} is a positive word when used as `touch my heart' in the movie domain, but it is a neutral word in the computer domain when used as `touch pad'). How to extract transferable knowledge jointly considering the source and target domains is the art of transfer learning.
In the literature, there are several concepts that are closely related to transfer learning hence mislead the readers, e.g., sample selection bias, covariate shift, class imbalance, domain adaptation, and multi-task learning. The research in \cite{5288526} tried to distinguish and organize them according to the settings of the source and target domains, e.g., labeled data exists or not in target domain. This paper does not explicitly distinguish these, however we consider all of them as transfer learning. Further discussion of these concepts and their distinctions can be found in \cite{5288526,weiss2016survey}. 
The capability of recognizing, modeling and utilizing transferable knowledge between two domains does not only improve the performances of specific real-world problems, it is also a significant step in the promotion of the self-learning (like humans) of robots without any human intervention. Imagine the following scenario: an intelligent robot faces a new problem about which it has no knowledge, so it asks for help from other robots within similar domains and learns from them, whereupon the problem is resolved. Therefore, we believe transfer learning has a promising future not only in the  statistical machine learning area but also for robots and even the general artificial intelligence.

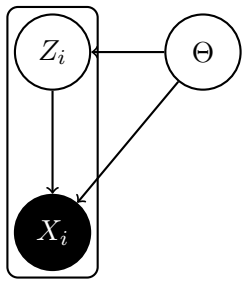
\begin{figure}
\centering
\begin{tikzpicture}[scale=0.4,
    fact/.style={circle, minimum size=1cm, draw=black, thick, text centered, fill=black, text=white},
    state/.style={circle, minimum size=1cm, draw=black, thick, text centered, text=black}
]
\draw[thick, rounded corners] (-1.5,1.5) rectangle (1.5,-7.5);
\node (State0) at (0,0) [state] {$Z_i$} ;
\node (State1) at (0,-6) [fact] {$X_i$} ;
\node (State2) at (5,0) [state] {$\Theta$} ;

\path [->, thick]	

(State0)    edge (State1)
(State2)    edge (State0)
(State2)    edge (State1)

;
\end{tikzpicture}
\caption{A simple probabilistic graphical model where the dark node $X_i$ denotes an observed datum and nodes $Z_i$ and $\Theta$ are two (local and global, respectively) latent variables}
\label{fig:pgm}
\end{figure}

The probabilistic graphical model (PGM) \cite{wainwright2008graphical,Koller+Friedman:09}, which is a significant branch of the statistical machine learning, is a rich framework for modeling (expressing) the complex interaction between a finite or infinite number of (observable or latent) variables from domains through probabilistic distributions or stochastic processes. The name comes from its structure - a graph with random variables as nodes and probabilistic dependencies as edges, as shown in Fig. \ref{fig:pgm}.
According to the edge type (i.e., directed or undirected) between nodes/variables, there are two categories of probabilistic graphical models: directed  and undirected. For example, hidden Markov models \cite{rabiner1989tutorial} are a type of directed graphical model; Conditional Random Field \cite{Lafferty:2001:CRF} is an example of an undirected graphical model.
Applying probabilistic graphical models to a target task comprises the following two steps: 1) model design and 2) model inference.
Given a task, the first step is to analyze the nature of the problem and then design a number of variables and their relationships to capture this nature. In other words, this step is to design the graphical structure of PGM which should jointly consider the observed data and additional knowledge of the target task. Note that there is no exact procedure for this step because it strongly depends on the view/understanding of different people working on the same problem. For example, in the Latent Dirichlet Allocation model \cite{blei2003latent}, the documents are modeled by random variables satisfying Dirichlet or multinomial distributions, and variables are linked by the Dirichlet-multinomial relationship; in the Gamma-Poisson model \cite{ogura2013gamma}, the documents are modeled by random variables satisfying Gamma or Poisson distributions, and variables are linked by the Gamma-Poisson relationship. It is normally difficult and meaningless to discuss the advantages and disadvantages without considering the specific task.
The output of PGM is the interested marginal or joint posterior distributions defined by the graphical model given the observed data. Also, the PGM from the first step is actually a family of models, because the designed probabilistic distributions are normally with unknown parameters and different parameter settings would lead to different models. With observed data in hand (some variables/nodes in graphical models are with known values), the second step is to infer posterior distributions of the latent variables and to estimate model parameters. For some sparse graphs, there is an exact algorithm to learn the PGM: the junction tree algorithm \cite{kahle2008junction,wainwright2008graphical}. Unfortunately, this algorithm is not applicable for complex graphical models which have many complicated tasks. Therefore, some approximation algorithms have been developed to resolve this problem: Expectation Maximization \cite{dempster1977maximum}, Laplace Approximations, Expectation propagation \cite{minka2001expectation}, Monte Carlo Markov Chain \cite{neal1993probabilistic}, and Variational Inference \cite{blei2017variational}. Furthermore, the designed probabilistic dependencies between variables may also not be fixed but learned from data (so-called \emph{structure learning}). An example is the Bayesian Network, where the network structure (i.e., dependencies between variables) can be learned from data. Due to their powerful modeling ability and solid theoretical foundations, probabilistic graphical models have attracted significant attentions from researchers within a variety of areas, e.g., molecular biology \cite{Friedman799}, text mining \cite{blei2003latent}, natural language processing \cite{SultanBS16}, and computer vision \cite{gupta2012bayesian}.

In comparison to other models (e.g., support vector machines) in machine learning, probabilistic graphical models have the following advantages that are potentially good for transfer learning: 1) handling uncertainty. Uncertainty appears in almost any real-world problems and, of course, their observations (data). For example, people may use different words when they write documents on a specific topic, so we need to consider this uncertainty when building models to uncover hidden topics. PGMs have the ability to handle (model) this uncertainty well with the help of probability distributions or stochastic processes; 2) handling missing data. One typical example of missing data is from recommender systems where users only rate a limited number of items so the ratings on other items are missing. PGM can competently handle this through a latent variable design \cite{NIPS20134899}; 3) interpretability. A PGM comprises defined probabilistic distributions (or stochastic processes), so human experts can evaluate its semantics and properties or even incorporate their knowledge into the model. Through the structure of PGM, people can easily understand the problem and the domain; 4) generalization ability. Directed PGMs (also known as \emph{generative models}) have a good generalization ability to compare discriminative models, especially with limited number of data \cite{NIPS20012020}.
Although a number of excellent studies on transfer learning have been published in the literature, such as: comprehensive ones \cite{5288526,weiss2016survey}, applications, e.g., reinforcement learning \cite{Taylor:2009:TLR}, collaborative filtering \cite{Li:2011:CCF}, visual categorization \cite{6847217}, face and object recognition \cite{7078994}, speech and language processing \cite{7415532}, activity recognition \cite{Cook2013}, and methodology, e.g., computational intelligence \cite{Lu:2015:TLU}, there is no one specific to the work on using PGMs for transfer learning.
We reviews the majority of the work in this area, identify the basic methods used by existing PGMs for transfer learning, and lay the foundation for further study in this direction.
This paper provides researchers in the transfer learning field with an overview of the work on using PGMs, and promotes the use of PGMs as a transfer method. This paper also provides researchers in the PGM field with an overview of the existing successful applications of PGMs on transfer learning, and promotes the developments of PGMs specific to transfer problems.
This paper is written with the assumption that readers have a basic knowledge of transfer learning.

\begin{figure}[t]
\centering
\includegraphics[scale=0.55]{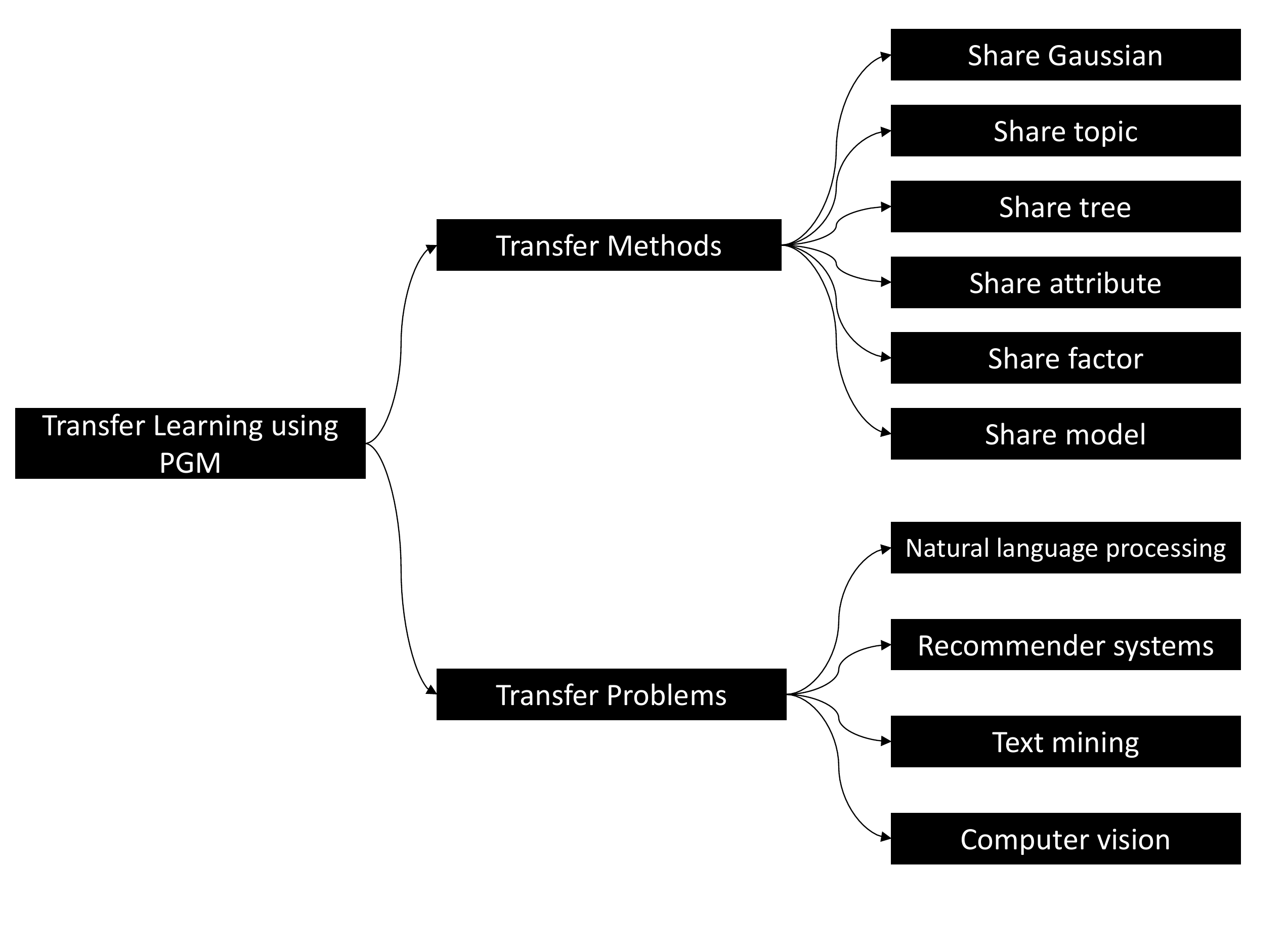}
\caption{The framework of the paper }
\label{fig:framework}
\end{figure}

The remaining part of this paper is structured as follows. Section 2 discusses the existing state-of-the-art methods used by probabilistic graphical models for transfer learning. Section 3 presents the real-world transfer learning problems solved by probabilistic graphical models. Finally, Section 4 concludes this paper and presents possible challenges for the further study.

\section{Transfer methods}

\begin{figure}
\centering
\begin{subfigure}{\textwidth}
\centering
  \includegraphics[scale=0.4]{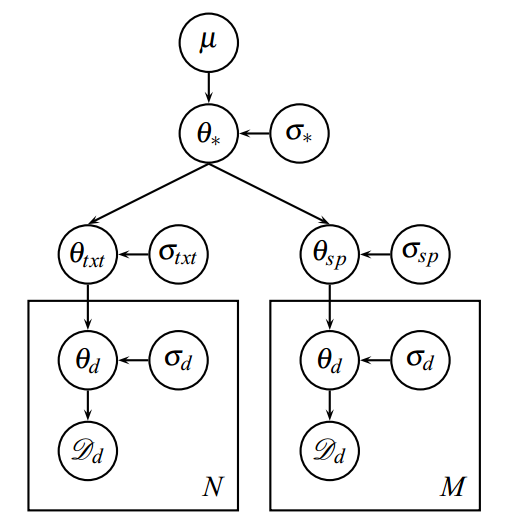}
\caption{A general hierarchical Bayesian domain adaptation framework \cite{FinkelM09a}}
\label{fig:model15}
\end{subfigure}
\begin{subfigure}{\textwidth}
\centering
  \includegraphics[scale=0.3]{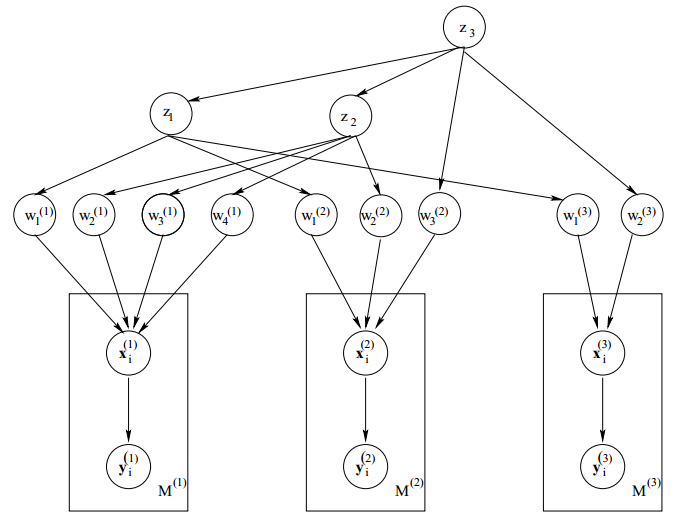}
\caption{Hierarchical Bayesian domain adaptation between CRFs \cite{ArnoldNC08}}
\label{fig:model16}
\end{subfigure}
\\
\begin{subfigure}{\textwidth}
\centering
  \includegraphics[scale=0.4]{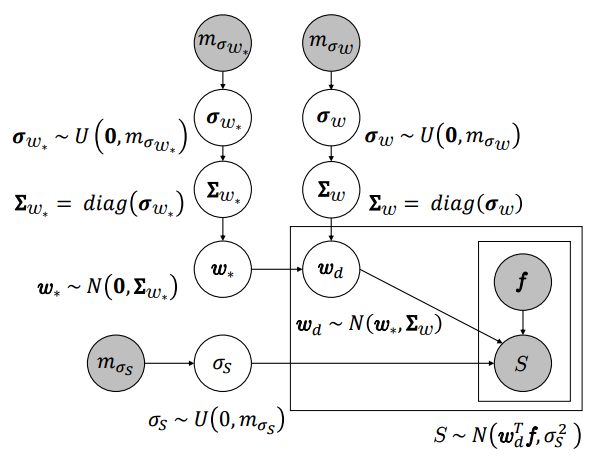}
\caption{Bayesian ridge/logistic regression models \cite{SultanBS16}}
\label{fig:SultanBS16}
\end{subfigure}
\begin{subfigure}{\textwidth}
\centering
  \includegraphics[scale=0.6]{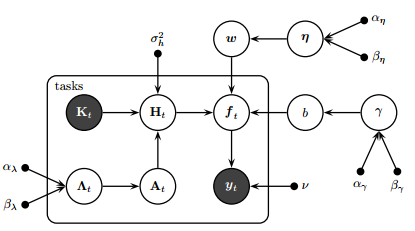}
\caption{Kernelized Bayesian transfer learning \cite{GonenM14}}
\label{fig:model17}
\end{subfigure}
\caption{Four illustrative PGMs for transfer learning using shared Gaussian mean}
\label{fig:gaussianmean}
\end{figure}

\begin{figure*}
\centering
\begin{subfigure}{\textwidth}
\centering
\includegraphics[scale=0.45]{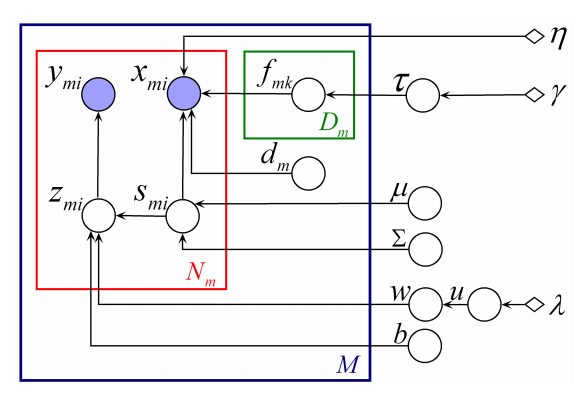}
\caption{ Latent Probit Model \cite{han2012cross} }
\label{fig:latentprobitmodel}
\end{subfigure}
\begin{subfigure}{\textwidth}
\centering
 \includegraphics[scale=0.45]{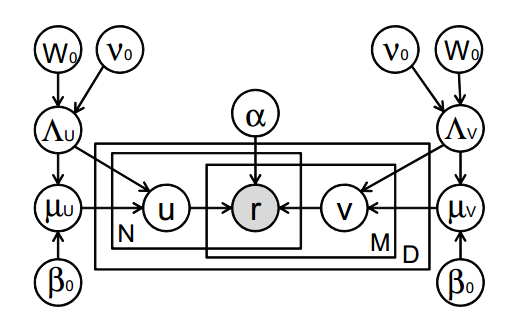}
\caption{Cross-domain recommendation without shared users or items \cite{Iwata2015} }
\label{fig:aistats2015}
\end{subfigure}
\\
\begin{subfigure}{\textwidth}
\centering
\includegraphics[scale=0.6]{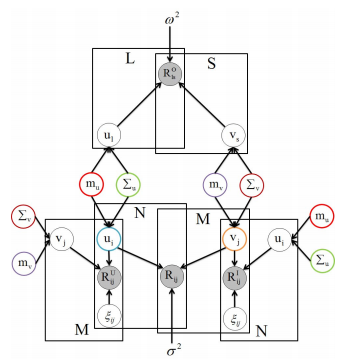}
\caption{Transfer Probabilistic Collective Factorization
Model \cite{7023342}}
\label{fig:7023342}
\end{subfigure}
\begin{subfigure}{\textwidth}
\centering
\begin{tikzpicture}[scale=0.45,
    fact/.style={circle, minimum size=1cm, draw=black, thick, text centered, fill=black, text=white},
    state/.style={circle, minimum size=1cm, draw=black, thick, text centered, text=black}
]
\draw[thick, rounded corners] (-1.5,1.5) rectangle (1.5,-7.5);
\node (State0) at (0,0) [state] {$U_i$} ;
\node (State1) at (0,-3) [fact] {$X_i$} ;
\node (State2) at (0,-6) [state] {$V_i$} ;

\node (State6) at (3,0) [state] {$\Omega$} ;

\draw[thick, rounded corners] (4.5,1.5) rectangle (7.5,-7.5);
\node (State3) at (6,0) [state] {$U_j$} ;
\node (State4) at (6,-3) [fact] {$X_j$} ;
\node (State5) at (6,-6) [state] {$V_j$} ;

\path [->, thick]	

(State0)    edge (State1)
(State2)    edge (State1)

(State6)    edge (State0)
(State6)    edge (State3)

(State3)    edge (State4)
(State5)    edge (State4)

;
\end{tikzpicture}
\caption{Multi-Domain Collaborative Filtering \cite{ZhangCY10} }
\label{fig:uai2010}
\end{subfigure}
\caption{Four illustrative PGMs for transfer learning using shared Gaussian mean and  variance. }
\label{fig:gaussianmeanvariance}
\end{figure*}

In this section, we review the designed transfer methods used by existing state-of-the-art PGMs for transfer learning.
The majority of research on transfer PGMs is based on the existing state-of-the-art PGMs originally developed for single domains, as summarized in Table \ref{basemodels}. When trying to bridge two different domains, the common idea of transfer PGMs is to share some statistical strengths between models for two domains. For example, additional random variables are introduced as a bridge to connect two domains and the knowledge from one domain could be transferred to the other domain through this variable. Once the model is fixed after the transfer method design, the inference is the same with traditional PGMs and existing inference algorithms could be adopted for this transfer PGM, so we do not discuss model inference in this paper. The design of a transfer bridge/variable strongly depends on the problem and the understanding of the researchers on this problem.
According to the differences of shared statistical strengths used in these works, we organize them into the following categories: 1) share Gaussian prior, 2) sharing topic, 3) sharing tree, 4) sharing attribute, 5) sharing factor, and 6) sharing model, as shown in Fig. \ref{fig:framework}.

\subsection{Share Gaussian prior}
\label{sharegaussianprior}

Gaussian distribution is a significant and well-studied probabilistic distribution on real-value space because it is often adopted when the underlying distributions are unknown.
The first group of works in this category is to share the mean of a Gaussian prior between source and target domains.
A general hierarchical Bayesian domain adaptation framework is proposed in \cite{FinkelM09a} shown in Fig. \ref{fig:model15}, where the model parameters in two domains are linked by a high-level variable satisfying a Gaussian distribution.
Another similar framework \cite{ArnoldNC08} is proposed with three layers, and each variable at higher layer is used as the mean value of the Gaussian prior for the lower layer variable as shown Fig. \ref{fig:model16}.
One instance of the above framework is the transfer Bayesian ridge/logistic regression model as shown in Fig. \ref{fig:SultanBS16}, where feature weighting vector $w_d$ for each domain is given a globally shared prior $w^*$ that is further given a zero-mean prior \cite{SultanBS16}. It is claimed that this model is the least affected by adverse factors (e.g., noisy training data) for Bayesian regression.
Another instance is the kernelized Bayesian transfer learning for binary/multi-class classification \cite{GonenM14} which is a fully Bayesian solution to domain adaptation on heterogeneous domain in the discriminative setting. As shown in Fig. \ref{fig:model17}, the bias term is also shared by domains except for the Gaussian prior for the feature weighting vectors $f$.
The second group of works in this category is to share the mean and covariance of a Gaussian prior between source and target domains. Sharing covariance actually encodes a stronger constraint for the two domains comparing the mean. 
The Bayesian ridge/logistic regression in \cite{SultanBS16} is also extended for a multi-task situation, and the feature weighting vectors $w_d$ for tasks are given a Gaussian prior with different means but the same variance. The means for tasks are further linked through a higher-level Gaussian prior which produces the byproduct of the model: task-level and domain-level feature weights.
A shared Gaussian prior is placed for weights of the radial basis function (RBF) network in a multi-task non-linear regression setting \cite{YangTD08}.
In the Latent Probit Model \cite{han2012cross}, the latent feature vectors for all data points in all domains are commonly drawn from a shared Gaussian distribution (both mean and variance). In order to take the domain-specific property into account, these latent feature vectors are additionally transformed by the domain-specific transforming vectors before generating the observed feature vectors and their corresponding labels as shown in Fig. \ref{fig:latentprobitmodel}. Similarly, the hidden variables for two factor matrices from the probabilistic matrix factorization \cite{mnih2008probabilistic} of the source domain are linked to the corresponding ones in the target domain by two shared Gaussian distributions (each one for a factor matrix) \cite{7023342,Iwata2015} in Figs. \ref{fig:7023342} and \ref{fig:aistats2015}. Another method proposed to link factor matrices from the probabilistic matrix factorization is to share a matrix-variate Gaussian distribution directly on factor matrices, e.g., the distribution is parameterized by a zero-mean and a covariance variable in \cite{ZhangCY10}.

\subsection{Share topic}
\label{sharetopic}

\begin{figure*}
\centering
\begin{subfigure}{0.45\textwidth}
\centering
\includegraphics[scale=0.4]{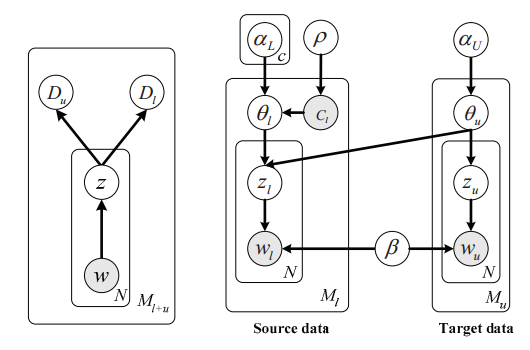}
\caption{Topic-bridged PLSA \cite{Xue:2008} and Topic-bridged LDA \cite{WuC10a}}
\label{fig:sharedtopics}
\end{subfigure}
\begin{subfigure}{0.45\textwidth}
\centering
\begin{tikzpicture}[scale=0.4,
    fact/.style={circle, minimum size=0.8cm, draw=black, thick, text centered, fill=black, text=white},
    state/.style={circle, minimum size=0.8cm, draw=black, thick, text centered, text=black}
]

\draw[thick, rounded corners] (-2,2) rectangle (2,-9.8);
\draw (1.5,-9.3) node[scale=0.7]  {$C$} ;
\draw[thick, rounded corners] (-1.5,1.5) rectangle (1.5,-1.3);
\draw (1,-1) node[scale=0.7]  {$K$} ;
\node (State0) at (0,0) [state] {$\theta_{j,i}$} ;

\draw[thick, rounded corners] (-1.8,-1.5) rectangle (1.8,-8.9);
\draw (1.2,-8.5) node[scale=0.7]  {$D$} ;
\draw[thick, rounded corners] (-1.5,-1.8) rectangle (1.5,-4.2);
\draw (1,-3.8) node[scale=0.7]  {$K$} ;
\node (State1) at (0,-3) [state] {$\pi_{d,j}$} ;

\draw[thick, rounded corners] (-1.5,-4.5) rectangle (1.5,-8);
\draw (1,-7.5) node[scale=0.7]  {$N_d$} ;
\node (State2) at (0,-6) [fact] {$w_{j,d,i}$} ;

\draw[thick, rounded corners] (4.5,1.5) rectangle (7.5,-1.5);

\node (State3) at (6,0) [state] {$\theta_k$} ;

\draw (7, -1) node[scale=1] {$K$};

\node (State4) at (6,-3) [state] {$\lambda_C$} ;
\node (State5) at (6,-6) [state] {$\theta_B$} ;
\node (State6) at (6,-9) [state] {$\lambda_B$} ;

\path [->, thick]	

(State0)    edge (State1)
(State1)    edge (State2)
(State3)    edge (State2)
(State4)    edge (State2)
(State5)    edge (State2)
(State6)    edge (State2)

;
\end{tikzpicture}
\caption{Cross-collection model \cite{Zhai:2004:CMM}}
\label{fig:KDD2004}
\end{subfigure}
\\
\begin{subfigure}{0.45\textwidth}
\centering
\includegraphics[scale=0.42]{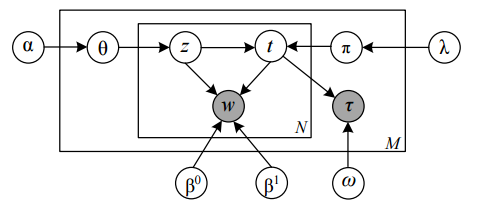}
\caption{$\tau$-LDA model \cite{yang2011bridging}}
\label{fig:languagegap}
\end{subfigure}
\begin{subfigure}{0.45\textwidth}
\centering
\includegraphics[scale=0.25]{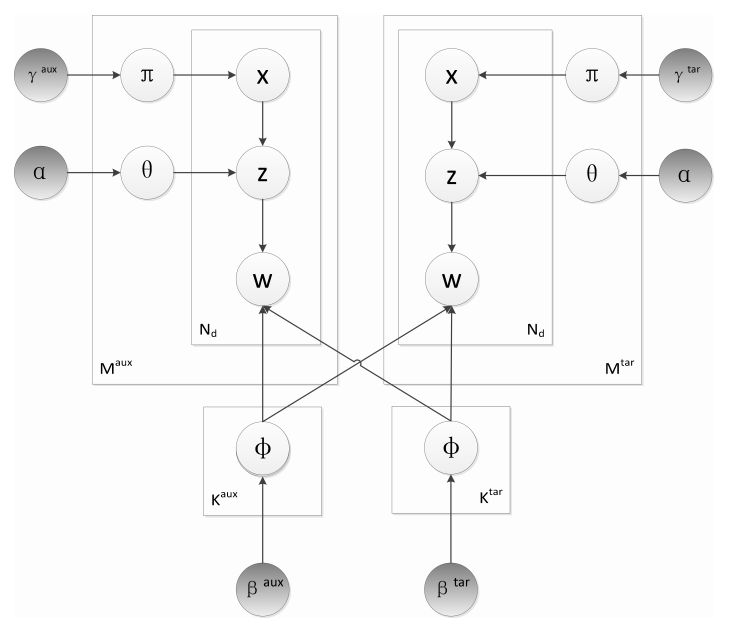}
\caption{Dual-LDA\cite{Jin:2011:TTK}}
\label{fig:DLDA}
\end{subfigure}
\\
\begin{subfigure}{0.45\textwidth}
\centering
\includegraphics[scale=0.25]{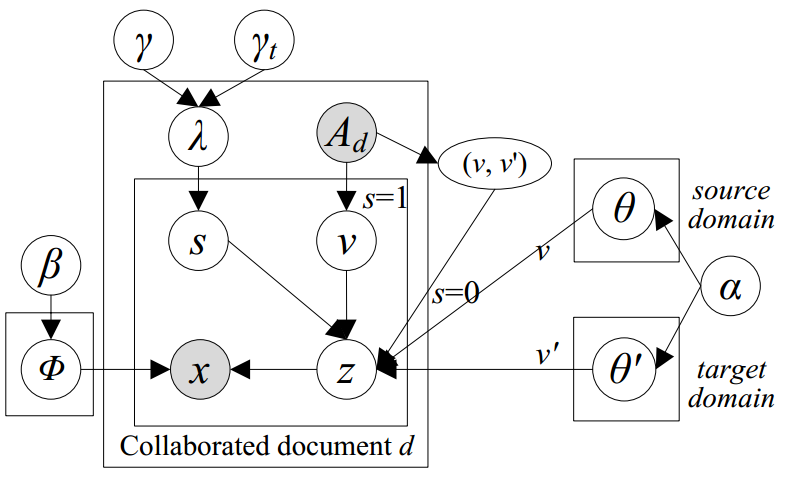}
\caption{Cross-domain topic learning\cite{Tang:2012:CCR}}
\label{fig:sharedtopicsbyctl}
\end{subfigure}
\begin{subfigure}{0.45\textwidth}
\centering
\includegraphics[scale=0.2]{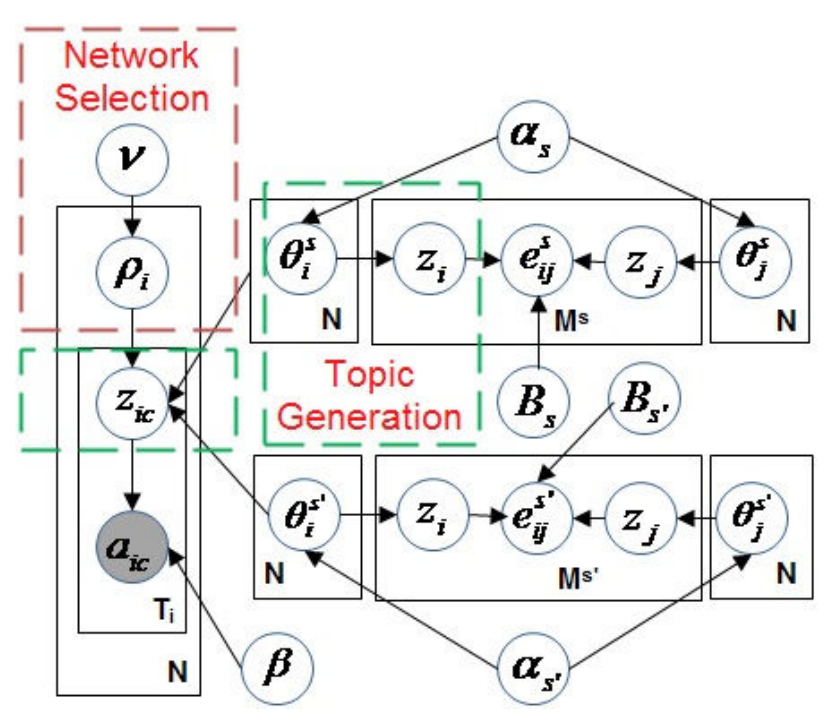}
\caption{Composite social topic model \cite{zhong2012comsoc}}
\label{fig:KDD20121}
\end{subfigure}
\\
\begin{subfigure}{0.45\textwidth}
\centering
\includegraphics[scale=0.3]{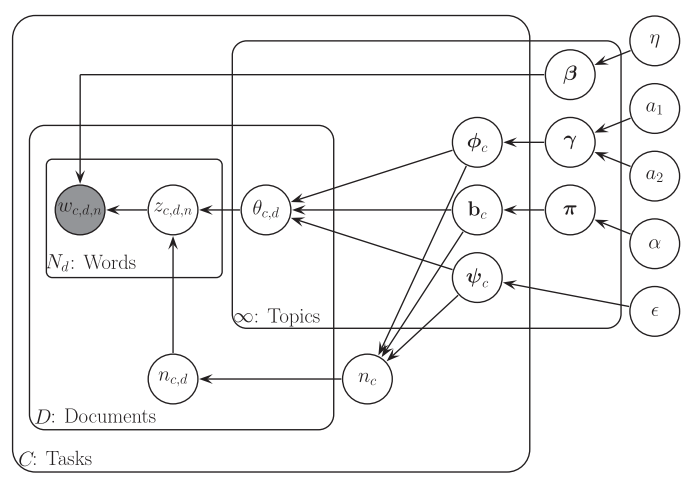}
\caption{Multi-task HDP \cite{Faisal2013124}}
\label{fig:sharedtopicsbyibp}
\end{subfigure}
\begin{subfigure}{0.45\textwidth}
\centering
\includegraphics[scale=0.3]{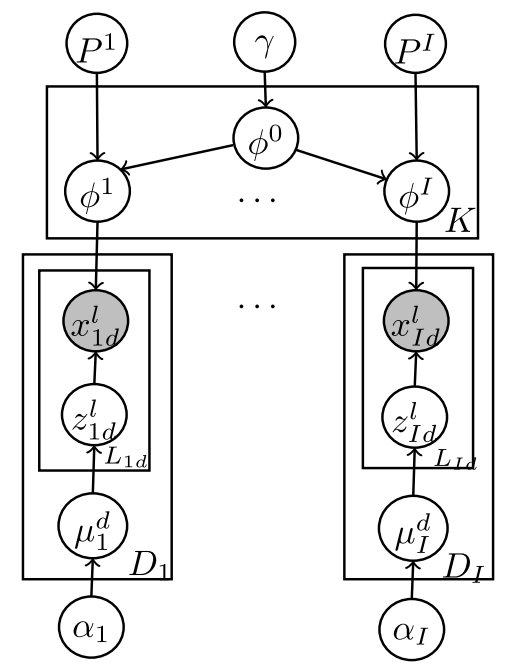}
\caption{Differential topic model \cite{ChenBDXD15}}
\label{fig:sharedtopicsbytpyp}
\end{subfigure}
\caption{Nine illustrative PGMs for transfer learning using shared topics based on PLSA, LDA, and Bayesian nonparametric models.}
\label{fig:topic}
\end{figure*}
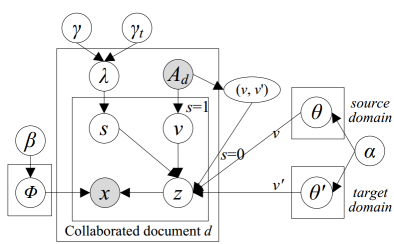
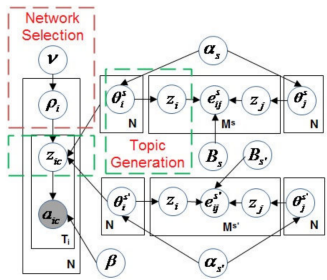
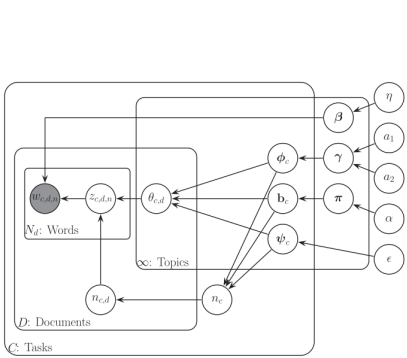
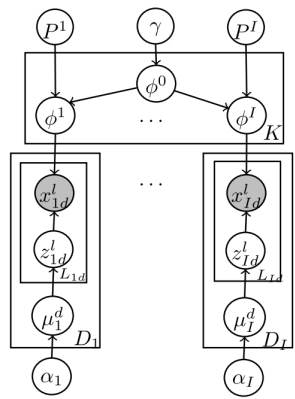

The word `topic' in PGMs might seem out of the text mining area which aims to model and understand documents. Probabilistic latent semantic analysis (PLSA) \cite{Hofmann:1999:PLS} and Latent Dirichlet allocation (LDA) \cite{blei2003latent} are commonly accepted as two of the earliest and most successful probabilistic graphical models for document modeling. They named a weighted (the weights on all dimensions are nonnegative and summation equals one) vector on the vocabulary of a corpus as a \emph{topic}, so the two models are also named \emph{topic models} \cite{Blei:2012:PTM}. For example, given a large number of research papers in the artificial intelligence area, topic models may discover several underlying topics from these papers, such as \emph{intelligent robotics}, \emph{natural language processing}, and \emph{computer vision}. Following their success, topic models have experienced an explosive growth in many areas, such as text mining, image processing, computer vision, social network, computational biology, etc.
With the application extension of various topic models, the meaning of topics has changed accordingly, e.g., a topic in image processing may denote \emph{people}, \emph{water}, or \emph{sky} \cite{Li:2005:BHM}. Therefore, the term \emph{topic} in this paper does not only mean a vector on vocabulary but refers to a vector with constraints (the weights on all dimensions are nonnegative and summation equals one).
Since the topics can be seen as frequent patterns summarized from the data and they can be seen as general underlying knowledge if the data is large enough, they may be helpful to task in another domain, so an idea is to share these topics between source domain(s) and the target domain.

The first group is based on probabilistic latent semantic analysis (PLSA) \cite{Hofmann:1999:PLS} which decomposes the joint probability of documents and words into three components, i.e., $P(d,w)=P(d)\sum_z P(w|z)p(z|d)$ or (equally) $P(d,w)=\sum_z P(z)P(w|z)p(d|z)$. Comparing traditional SVD-based latent semantic analysis \cite{ASI:ASI1}, PLSA has the following advantages: 1) components have clear probabilistic meanings, e.g., $P(w|z)$ are nonnegative and normalized vectors, and 2) it builds on the social statistical theory, e.g., it is easy to do model selection.
Topic-bridged PLSA \cite{Xue:2008} tries to link two domains by sharing the component $P(w|z)$ (i.e., topics) in Fig. \ref{fig:sharedtopics}. In order to fully utilize the knowledge of the source domain, additional penalty terms are added for \emph{must-link} and \emph{cannot-link} constraints.
An extension of PLSA (named Dual-PLSA \cite{4959893,GaoL11}) introduces another latent variable $\tilde{z}$ aiming to separate the documents and words because they may express different topics. Based on Dual-PLSA, a Collaborative Dual-PLSA model \cite{Zhuang:2010,5936065} is proposed with a latent variable for labels and dependencies on words and documents, but the $p(z, \tilde{z})$ which are independent of the label variable is shared across domains.
Also, the HIDC model \cite{ZhuangLYHS13} is proposed based on the Dual-PLSA to account for different types of topics (named concepts in \cite{ZhuangLYHS13}): identical topics, which are shared by domains; distinct topics, which only appear in one domain; and homogeneous topics, which are mediate.
Another extension of PLSA is to revise the word distribution $P(w|z)$ to word-pair distribution $P(w_1, w_2|z)$, and Cross-domain Topic Indexing (CDTI) \cite{GaoL11} is proposed to mine the topics shared by two domains, where topics are defined as word-pair distributions rather than word distributions which models the cross-domain word co-occurrence relations.
The aforementioned models all try to learn the shared topics between domains but the domain-specific properties are ignored, and irrelevant topics may degrade the task performance in the target domain. In order to learn shared and unique topics at the same time, the Joint Mixture Model (JMM) \cite{Li:2012} introduces a Bernoulli variable to explicitly separate two types of topics based on PLSA, and this model further evaluates the similarity between shared and unique topics by Jensen-Shannon divergence and the correlations between domain-specific topics using Pearson's Correlation Coefficients. The Cross-Collection Mixture Model \cite{Zhai:2004:CMM} is another model for shared (background/common) and unique topics, where normalized weights are used to control the selection of different types topics during the generative procedure in Fig. \ref{fig:KDD2004}. In \cite{li2009transfer}, user- and item-cluster level rating patterns, which are nonnegative and normalized vectors (called topics here), are shared by domains and generate ratings like PLSA.

The second group is based on latent Dirichlet allocation (LDA) \cite{blei2003latent}, which resolves the possible over-fitting problem of PLSA and has the ability to handle unseen documents by introducing a Dirichlet prior for all topics.
Similar to topic-bridged PLSA \cite{Xue:2008}, a topic-bridged LDA \cite{Xue:2008} is also developed by sharing topics between domains as shown in Fig. \ref{fig:sharedtopics}, and a category topic model (CT) \cite{YuA10} also shares topics between domains (multiple categories) by extending the author-topic model \cite{rosen2004author} using informative priors.
$\tau$-LDA model \cite{yang2011bridging} also shares topics between domains but it also considers the domain-specific property in Fig. \ref{fig:languagegap}. The topics in this model are composed of two sets that account for two extremes (i.e., expert and layman), and each word is generated by a selected topic (for the two extremes)  controlled by a topic-specific switching variable.
Different from above two models with shared topics between domains, dual-LDA \cite{Jin:2011:TTK} in Fig. \ref{fig:DLDA} explicitly separates the topics for two domains linked by a binary switch variable that controls the topic selection of words. This method is expected to automatically capture whether a document in the source domain should relate to target domain.
The same idea is also adopted by Cross-domain Topic Learning (CTL) \cite{Tang:2012:CCR} in Fig. \ref{fig:sharedtopicsbyctl} which firstly learns the topics from both source and target domains, and then learns the topics from collaborative papers of authors from both domains.
In Fig. \ref{fig:KDD20121}, the Composite Social Topic model (ComSoc) \cite{zhong2012comsoc} transfers user interest in topics in different networks to user-item interactions, where the user interest vectors in different social networks are linked by an assignment variable that controls the generation of user-item interactions.
As LDA belongs to unsupervised learning, other works uses LDA as a step followed by another learning model (e.g., Naive Bayes \cite{rish2001empirical}) rather than building a complete PGM by extending LDA. Frequent word-sets of topics from multi-domains are mined using LDA and are used as transferable knowledge to assist topic mining in target domains \cite{liu2014topic}, where the similarity between a current topic in the target domain with a topic from other domain is measured by symmetrised KL divergence; and topics learned from the source domain using LDA are directly used to construct new representations for documents from the target domain \cite{5416713}.
Although the existing work is successful in terms of their own tasks, it would be more interesting and practical to infer the shared topics between (source and target) domains and domain-specific topics for each domain with an additional constraint that the domain-specific topics of two domains should be as distinct from each other as possible. This constraint may help us remove irrelevant knowledge when it is transferred.

The third group is based on Bayesian nonparametric topic models. The work in the aforementioned two groups is based on PGMs built by probabilistic distributions (e.g., Gaussian, multinomial, Dirichlet, beta, Bernoulli distributions), and the work in this group is based on PGMs built by stochastic processes (e.g., Gaussian process (GP) and Dirichlet process (DP)) which are also named Bayesian nonparametric models \cite{GERSHMAN20121,orbanz2011bayesian}. The advantage of Bayesian nonparametric models is their ability to work on an infinite number of topics and it allows the data itself determine the final topic number.
In Clustered Naive Bayes \cite{roy2007efficient}, a number of Naive Bayes share topics (named parameters in \cite{roy2007efficient}) from DP, where the tasks within the same group share the same topic and the well-trained task can help to train the insufficiently-trained task.
An ingenious extension of DP is to stack them hierarchically, i.e., hierarchical Dirichlet processes (HDP) \cite{hdp2006}. With the help of the discrete nature of random measures from DP, the topics/components from the global DP could be shared by the local DPs in HDP and thus it could benefit the transfer learning problem. One successful application of HDP in transfer learning is \cite{CaniniSG10} where topics/clusters are shared by categories.
Different from HDP, the activated topics are controlled by an additional Indian Buffet Process (IBP) \cite{griffiths2011indian} in Fig. \ref{fig:sharedtopicsbyibp}, and its advantage over HDP is that the sharing of topics and their weights could be decoupled which makes the sharing of low-weight topics possible between domains \cite{Faisal2013124}.
The Pitman-Yor process (PYP) \cite{pitman1997} is an two-parameter extension of DP with an additional property that the weights in a random measure from PYP satisfy a power-law distribution. It is natural to observe the power-law property in real life, e.g., word frequencies in natural languages, so PYP is normally adopted for natural language processing. Hierarchal PYP \cite{teh2006hierarchical} is also able to benefit the transfer learning analogy to HDP. For example, a set of basic topics is firstly generated for the sharing, and each domain uses these topics as the base measure of a Transformed PYP which transforms these topics to another set of topics using a domain-specific transformation matrix \cite{ChenBDXD15}, as shown in Fig. \ref{fig:sharedtopicsbytpyp}.
We believe that Bayesian nonparametric models have great potential for transfer learning because the scale of transferable knowledge between two domains is normally unknown and Bayesian nonparametric models are extremely flexible and can model this flexible transferable knowledge.
Note that we also found other work which uses Bayesian nonparametric topic models for transfer learning, but they all transfer knowledge between domains by building a hidden tree, so we leave this work to the following subsection.

\subsection{Share tree}
\label{sharetree}

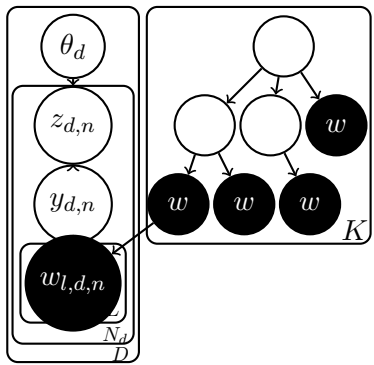
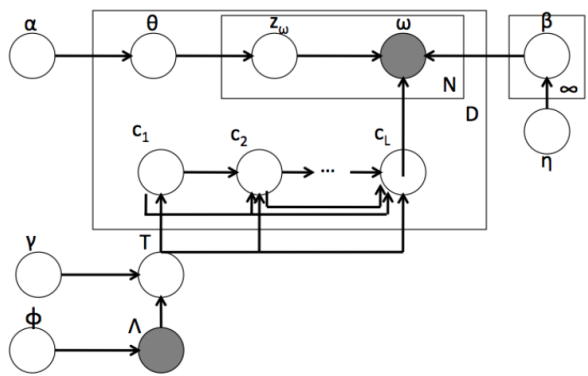
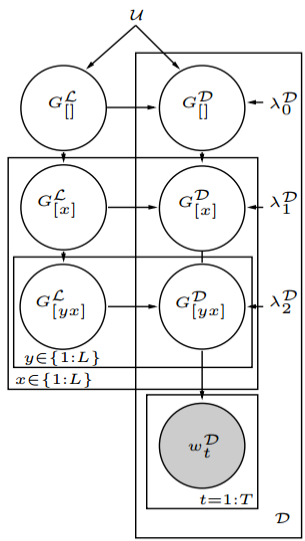
\begin{figure*}
\centering
\begin{subfigure}{\textwidth}
\centering
\begin{tikzpicture}[scale=0.35,
    fact/.style={circle, minimum size=0.8cm, draw=black, thick, text centered, fill=black, text=white},
    state/.style={circle, minimum size=0.8cm, draw=black, thick, text centered, text=black}
]

\draw[thick, rounded corners] (-2.5,1.5) rectangle (2.5,-12);
\draw (1.8,-11.7) node[scale=0.7]  {$D$} ;
\draw[thick, rounded corners] (-2.3,-1.5) rectangle (2.3,-11.3);
\draw (1.6,-11) node[scale=0.7]  {$N_d$} ;
\draw[thick, rounded corners] (-2,-7.5) rectangle (2,-10.5);
\draw (1.5,-10) node[scale=0.7]  {$L$} ;
\node (State0) at (0,0) [state] {$\theta_{d}$} ;
\node (State1) at (0,-3) [state] {$z_{d,n}$} ;
\node (State2) at (0,-6) [state] {$y_{d,n}$} ;
\node (State3) at (0,-9) [fact] {$w_{l,d,n}$} ;

\draw[thick, rounded corners] (2.8,1.5) rectangle (11.3,-7.5);
\draw (10.7, -7) node[scale=1] {$K$};

\node (State4) at (8,0) [state] {} ;

\node (State5) at (5,-3) [state] {} ;
\node (State6) at (7.5,-3) [state] {} ;
\node (State7) at (10,-3) [fact] {$w$} ;

\node (State8) at (4,-6) [fact] {$w$} ;
\node (State9) at (6.5,-6) [fact] {$w$} ;
\node (State10) at (9,-6) [fact] {$w$} ;

\path [->, thick]	

(State0)    edge (State1)
(State1)    edge (State2)
(State2)    edge (State3)

(State4)    edge (State5)
(State4)    edge (State6)
(State4)    edge (State7)

(State5)    edge (State8)
(State5)    edge (State9)
(State6)    edge (State10)

(State8)    edge (State3)
;
\end{tikzpicture}
\caption{Polylingual tree-based topic model \cite{HuZEB14}}
\label{fig:ACL2014}
\end{subfigure}
\begin{subfigure}{\textwidth}
\centering
\includegraphics[scale=0.4]{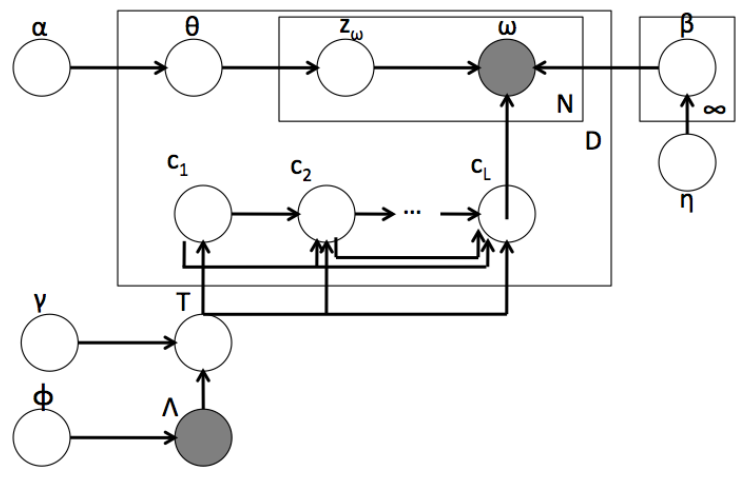}
\caption{Transfer hLDA \cite{KangML12}}
\label{fig:sharedtopictree}
\end{subfigure}
\begin{subfigure}{\textwidth}
\centering
\includegraphics[scale=0.5]{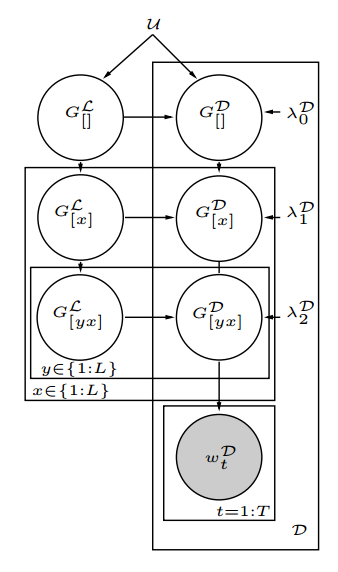}
\caption{Doubly HPYLM \cite{WoodT09}}
\label{fig:sharedhpylm}
\end{subfigure}
\caption{Three illustrative PGMs for transfer learning using a shared tree structure.}
\label{fig:tree}
\end{figure*}

Knowledge has inherent multi-granularity. Take the artificial intelligence research as an example again. This area comprises several research fields: \emph{intelligent robotics}, \emph{natural language processing}, \emph{computer vision}, etc. \emph{Natural language processing} further comprises several research directions: \emph{machine translation}, \emph{named-entity recognition}, \emph{question \& answering}, etc., and \emph{computer vision} comprises several research directions: \emph{image classification}, \emph{human activity recognition}, \emph{object tracking}, etc. All these research directions comprises several research points as well. This can be illustrated as a tree structure where 1) the root node denotes \emph{artificial intelligence}; 2) the father node is more general than its child node, 3) the child node is a specific aspect of its father node; 4) and the node number at the higher layer is usually smaller than the one at lower layer because a father node is more general to cover its child nodes.
Since transferable knowledge is special knowledge, it is safe to claim that the transferable knowledge between two domains also has multi-granularity. This claim is useful for transfer learning because usually, the more general the knowledge in the source domain, the larger the probability it will be  reusable in a (related) target domain compared to specific knowledge. For example, general knowledge of \emph{horse}, \emph{cows}, and \emph{sheep} could be used to understand the behavior of \emph{wildebeest}, but specific knowledge of \emph{sheep} may not reusable for \emph{wildebeest} \cite{SalakhutdinovTT12}.
When we want to facilitate transfer learning with the help of a tree, there are two problems that should be resolved: 1) how to reasonably build/mine this tree? 2) how to appropriately use this tree in the target domain?
This section summarizes the research on PGMs which transfer knowledge between domains using tree structures.

Depending on whether the tree structure is learned or pre-provided, we categorize the work into two groups. In the first group, the tree used to assist transfer learning is already ready for use.
One example is the transfer Hierarchical LDA (thLDA) \cite{KangML12}, as shown in Fig. \ref{fig:sharedtopictree}, which transfers the knowledge in an existing tree (i.e., a topic hierarchy) to the target domain, where a path of the tree from the root to the bottom node is sampled for each document in the target domain. The nested Chinese Restaurant process (nCRP) \cite{Blei:2010:NCR}, which is a stochastic process that can assign probability to the path of a tree with infinite depth and width, is modified as the prior for paths. Another example is the polylingual tree-based topic model \cite{HuZEB14}, shown in Fig. \ref{fig:ACL2014}, which uses a similar method with nCRP to use a tree by assigning the path a probability that is a product of the weights of branches in this path, where the tree is also separately built from the source domain.
The second group learns a latent tree together with the knowledge transfer process. The above nCRP is not only a good tool for operating a tree structure but also a Bayesian nonparametric model to learn a latent tree by nesting the stochastic processes (i.e., CRPs). In addition to thLDA, the second example is \cite{SalakhutdinovTT12} where a tree structure is built with a fixed three-layer depth and unbounded width through nCRP. In this model, domains are treated as nodes at the bottom layer (named `level 1'), the parent nodes of the domains are named \emph{supercategory} at the middle layer (named `level 2'), and the root node at the topic layer (named `level 3') is a set of two variables. Each domain is characterized by a Gaussian distribution with a mean and a variance, and domains are expected to share similar parameters with their parent nodes (i.e., supercategories or root nodes).
The third example treats the parameters in the tree as transferable knowledge to assist the classification of classes with less labeled data \cite{NIPS2013_5029} where a tree structure with leaf nodes as labels/domains is built with nCRP prior.
One exception is the doubly Hierarchal Pitman-Yor language model \cite{WoodT09,teh2006hierarchical}, as shown in Fig. \ref{fig:sharedhpylm}, which builds a latent HPYLM as the transferable knowledge and, simultaneously, two separate HPYLMs are built for the two domains with the latent HPYLM as part of the base measures in PYPs in two HPYLMs.
In fact, the nested HDP (nHDP) \cite{6802355} based on nCRP was recently proposed as a prior for the subtrees of a tree, and it provides a more flexible way to operate the latent tree. We expect to see work which uses this new tree operation being developed in more flexible and ingenious ways to build or operate the latent trees for transfer learning within PGMs.

\subsection{Share attribute}
\label{shareattribute}

\begin{figure*}
\centering
\begin{subfigure}{0.4\textwidth}
\centering
\includegraphics[scale=0.26]{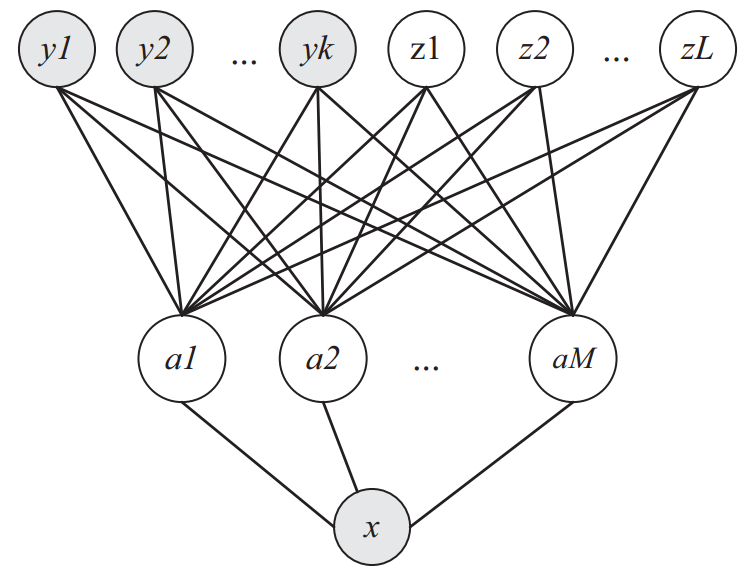}
\caption{DAP}
\label{fig:dap}
\end{subfigure}
\begin{subfigure}{0.4\textwidth}
\centering
\includegraphics[scale=0.25]{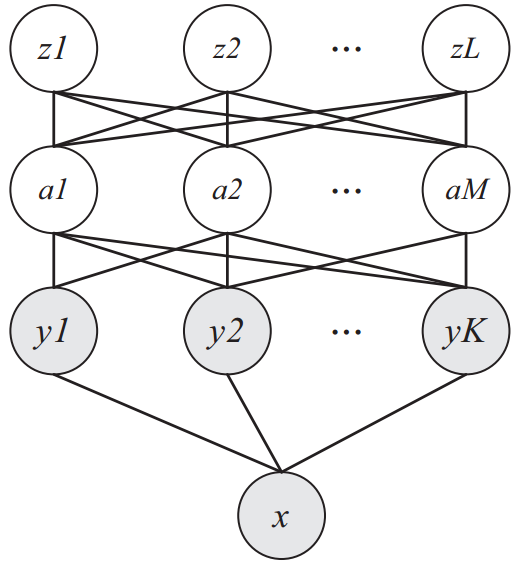}
\caption{IAP}
\label{fig:iap}
\end{subfigure}
\\
\begin{subfigure}{0.4\textwidth}
\centering
\includegraphics[scale=0.42]{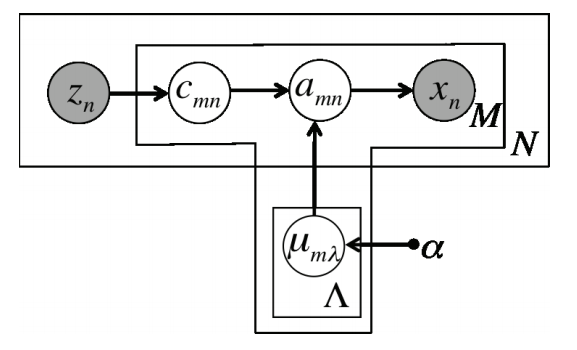}
\caption{Generative DAP model \cite{SuzukiSOK14}}
\label{fig:smc2014}
\end{subfigure}
\begin{subfigure}{0.4\textwidth}
\centering
\includegraphics[scale=0.5]{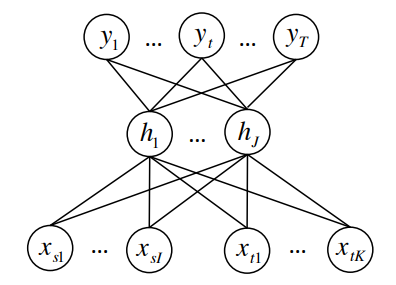}
\caption{RBM-based heterogeneous transfer learning \cite{WeiHTL2011}}
\label{fig:aaai2011}
\end{subfigure}
\caption{Four illustrative PGMs for transfer learning using shared attributes.}
\label{fig:attribute}
\end{figure*}

Using attributes to describe objects was firstly proposed in \cite{5206772}, which leads to further attribute-based transfer learning \cite{lampert2009learning,6571196}. A definition given in \cite{6571196} is: ``\emph{We call a property of an object an attribute, if a human has the ability to decide whether the property is present or not for a certain object.}''
Attributes, which are deemed as high-level descriptions about properties of objects, are believed to express more semantics than raw features, so they can be shared between object categories. For example, \emph{color}, \emph{texture}, \emph{shape}, and \emph{part} are attributes of images; \emph{has-horn}, \emph{has-leg}, \emph{has-head}, and \emph{has-wool} are attributes used for object categorization \cite{5206772}. In practice, these attributes are assumed to be provided by human experts, which is not expected to take much manual effort because the scale of the attributes is the same as one of the object classes. Although attributes look similar to topics discussed in Section \ref{sharetopic}, they are different because attributes are binary (and of course unnormalized) variables but topics are defined as nonnegative and normalized vectors in this paper.
As shown in Figs. \ref{fig:dap} and \ref{fig:iap}, direct attribute prediction (DAP) and indirect attribute prediction (IAP) \cite{5206594,6571196,LiZZZX13} are the first attribute-based transfer learning models, where the mapping relationship between classes and features is shared between the object classes. However, these two models are more like a framework and have no explicit and rigid probabilistic setup. The first directed PGM based on them is proposed in \cite{SuzukiSOK14}, shown in Fig. \ref{fig:smc2014}, which uses observation probability as a prior of attributes and additionally considers the attribute frequencies in the raw features. This directed PGM is for homogeneous transfer learning, and another attribute-based undirected PGM for heterogeneous transfer learning is shown in Fig. \ref{fig:aaai2011}, which is based on the restricted Boltzmann machine (RBM) \cite{smolensky1986information}. A typical RBM is composed of binary-valued hidden variables (here, named attributes) and visible units (two data features), and it defines a joint probability of attributes and data with a component (named energy) parameterized by the configurations of attributes. The attributes in this model are used to couple two feature representations in two domains.

\subsection{Share factor}
\label{sharefactor}

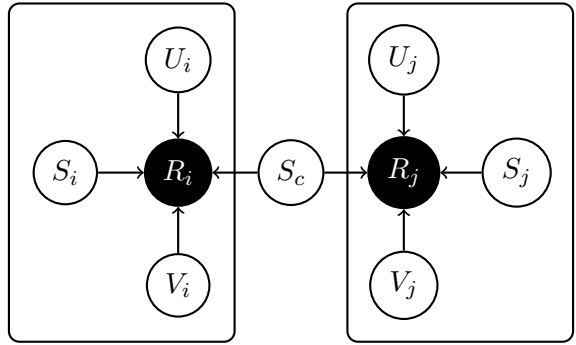
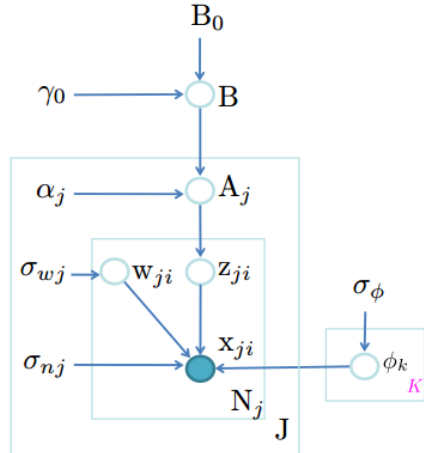
\begin{figure*}
\centering
\begin{subfigure}{0.5\textwidth}
\centering
\begin{tikzpicture}[scale=0.5,
    fact/.style={circle, minimum size=0.8cm, draw=black, thick, text centered, fill=black, text=white},
    state/.style={circle, minimum size=0.8cm, draw=black, thick, text centered, text=black}
]

\draw[thick, rounded corners] (-4.5,1.5) rectangle (1.5,-7.5);
\node (State0) at (0,0) [state] {$U_i$} ;
\node (State1) at (0,-3) [fact] {$R_i$} ;
\node (State2) at (0,-6) [state] {$V_i$} ;
\node (State3) at (-3,-3) [state] {$S_i$} ;

\node (State9) at (3,-3) [state] {$S_c$} ;

\draw[thick, rounded corners] (4.5,1.5) rectangle (10.5,-7.5);
\node (State5) at (6,0) [state] {$U_j$} ;
\node (State6) at (6,-3) [fact] {$R_j$};
\node (State7) at (6,-6) [state] {$V_j$} ;
\node (State8) at (9,-3) [state] {$S_j$};

\path [->, thick]	

(State0)    edge (State1)
(State2)    edge (State1)
(State3)    edge (State1)

(State5)    edge (State6)
(State7)    edge (State6)
(State8)    edge (State6)

(State9)    edge (State1)
(State9)    edge (State6)

;
\end{tikzpicture}
\caption{Probabilistic cluster-level latent factor model \cite{RenGLG15}}
\label{fig:aaai2015}
\end{subfigure}
\begin{subfigure}{\textwidth}
\centering
\includegraphics[scale=0.6]{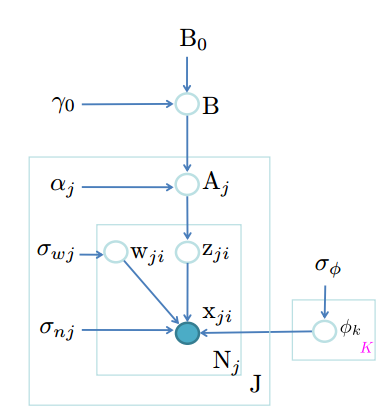}
\caption{HBP-based NJFA \cite{gupta2012bayesian}}
\label{fig:sharedfactors}
\end{subfigure}
\caption{Two illustrative PGMs for transfer learning using shared factor.}
\label{fig:factor}
\end{figure*}

Factor analysis is to describe or capture the variability of observed or correlated data with the help of a collection of unobserved variables called factors \cite{thompson2004exploratory}.
A mathematical definition is $Y=\Phi X+E$, where $Y$ is data, $\Phi$ is factors, $X$ is (factor) loading matrix, and $E$ is error. Similar to \emph{topics} detailed in Section \ref{sharetopic} and \emph{attributes} detailed in \ref{shareattribute}, factors can also be seen as high-level semantic descriptions of data, so they can also be transferred from one domain to another domain. Different from \emph{topics} in Section \ref{sharetopic} and \emph{attributes} in \ref{shareattribute}, factors are  real-valued vectors without additional constraints, i.e., nonnegativity and normalization of \emph{topics} and binary of \emph{attributes}. Although  traditional factor analysis for transfer learning has been extensively studied, there are a limited number of studies on PGM-based factor analysis for transfer learning.
One example of a factor-based transfer PGM is the Probabilistic Cluster-level Latent Factor Model \cite{RenGLG15}, shown in Fig. \ref{fig:aaai2015}. The factors in this model denote rating patterns, i.e., the probabilities of ratings from user clusters on item clusters, and these patterns are further decomposed into two parts: common rating patterns $S_{com}$ and domain-specific rating patterns $S_{spez}$.
Another example is Bayesian nonparametric joint factor analysis (NJFA) \cite{gupta2012bayesian}, shown in Fig. \ref{fig:sharedfactors}. In this model, domains are jointly factorized with shared factors and domain-specified factors. Different from the simply sharing factors between two domains by the matrix product and summation, the factor sharing in NJFA is more ingenious through using the hierarchical Beta process (HBP) \cite{thibaux2007hierarchical} prior. HBP is used to control the assignment of factors between domains, and its advantages are: 1) factors are shared by domains through hieratically dependent Beta processes; 2) the number of both shared and domain-specified factors does not need to be prefixed and can be automatically learned from the data.

\subsection{Share model}

All aforementioned models try to separately model the data from different domains and knowledge transfer is facilitated through the shared statistical strengths, e.g., topic, tree, and attribute. Other research uses the same model for both source and target domains and knowledge transfer is facilitated based on the generalizability of the probability models themselves.
One example is classification probabilistic principal component analysis (CPPCA) \cite{ChengL11} based on PPCA \cite{Yu:2006:SPP}, i.e., $X = W Z + \mu_x$, where the labeled training data and unlabeled test data are fed to the model together although they are from different domains with different covariances. CPPCA has the ability to model these covariances through the latent variables $Z$.
Another example is the Bayesian adaptation \cite{VillalbaL12} of probabilistic linear discriminant analysis (PLDA) \cite{prince2007probabilistic} which use the posterior distribution trained on data from the source domain as the prior of the target domain. There are no variables or parameters specially designed for one domain, and transferable knowledge is encoded in this posterior-prior distribution.


\begin{table*}
\renewcommand{\arraystretch}{1.2}
\caption{A summary of classical probabilistic graphical models extended for transfer learning}
\centering
\begin{tabular}{ p{7cm} p{0.21cm} p{7cm}   }
\hline
\textbf{Probabilistic graphical model}  & & \textbf{Used for transfer learning}
\\ \hline
Bayesian linear and logistic regression \cite{friedman2001elements}
&& \cite{SultanBS16}
\\ 
Probabilistic matrix factorization model (PMF) \cite{mnih2008probabilistic}
&& \cite{7023342} \cite{Iwata2015}
\\ 
Flexible mixture model \cite{si2003flexible}
&& \cite{li2009transfer}
\\ 
Polylingual topic models \cite{Mimno:2009}
&&\cite{HuZEB14}
\\ 
Probabilistic latent semantic analysis (PLSA) \cite{Hofmann:1999:PLS}
&&\cite{Xue:2008},\cite{GaoL11}, \cite{ZhuangLYHS13}, \cite{Zhuang:2010}, \cite{5936065}, \cite{Li:2012}, \cite{Zhai:2004:CMM}, \cite{li2009transfer}
\\ 
Latent Dirichlet allocation (LDA) \cite{blei2003latent}
&& \cite{WuC10a}, \cite{Jin:2011:TTK}, \cite{ChenBDXD15}, \cite{YuA10}, \cite{yang2011bridging}, \cite{Tang:2012:CCR}, \cite{5416713}
\\ 
Probabilistic linear discriminant analysis (PLDA) \cite{prince2007probabilistic}
&& \cite{7472720} \cite{VillalbaL12}
\\ 
Conditional random field (CRF) \cite{Lafferty:2001:CRF}
&& \cite{nallapati2010blind} \cite{FinkelM09a} \cite{ArnoldNC08}
\\ 
Hierarchal latent Dirichlet allocation (hLDA) \cite{blei2010nested}
&& \cite{KangML12}
\\ 
Pitman-Yor process mixture model \cite{teh2006hierarchical}
&& \cite{WoodT09} \cite{ChenBDXD15}
\\ 
Dirichlet process mixture model
&& \cite{Faisal2013124} \cite{roy2007efficient} \cite{CaniniSG10} \cite{hdp2006}
\\ 
Probabilistic principal component analysis (PPCA) \cite{Yu:2006:SPP}
&& \cite{ChengL11}
\\ 
Probit model \cite{muthen1979structural}
&& \cite{han2012cross}
\\
Nested Chinese restaurant process (nCRP) \cite{Blei:2010:NCR}
&& \cite{KangML12} \cite{SalakhutdinovTT12} \cite{NIPS2013_5029}
\\
Restricted Boltzmann machine (RBM) \cite{smolensky1986information}
&& \cite{SuzukiSOK14}
\\ \hline
\end{tabular}
\label{basemodels}
\end{table*}

\section{Transfer problems}
\label{applications}

This section reviews the real-world transfer problems solved by PGMs.
According to the application areas to which these transfer problems belong, we divide them into the following: \emph{Natural language processing}, \emph{Recommender systems}, \emph{Text mining}, \emph{Computer vision}, and \emph{Others}, as shown in Fig. \ref{fig:framework}.
These transfer problems demonstrate that PGMs have already been adopted in various areas. However, there is still a huge distance between the number of problems solved by transfer PGMs and the large number of problems solved by traditional single-domain PGMs \cite{Koller+Friedman:09}. We believe this distance is the potential for further development of PGMs for more real-world transfer problems. It is expected that this paper will motivate researchers from different areas to apply PGMs to more diverse transfer problems through the following examples.

\subsection{Natural language processing}

Natural language processing (NLP) \cite{manning1999foundations} is a research area focusing on the modeling, understanding, and utilization of  human languages. A typical example of this area is machine translation which aims to automatically translate one language (e.g., English) into another language (e.g., Chinese). Since the late 1980s, machine learning models or algorithms have dominated the NLP area \cite{hutchins2007machine,manning1999foundations}, and also a large number of PGMs have been developed for NLP. For example, Short Text Similarity (STS) evaluates the semantic similarity between two short text snippets (target domain). Since they are short (only a few sentences, e.g., a tweet), standard statistical methods do not work well, so additional information (source domain) like news webpages is introduced to facilitate this task \cite{SultanBS16}. Statistical machine translation (SMT) automatically translates sentences from one language (source domain) into another language (target domain) by learning from millions of former translations \cite{HuZEB14,WoodT09}. Named-entity recognition (NER) accurately labels the documents with named-entity tags. With the help of news webpages (source domain), PGMs \cite{ArnoldNC08,FinkelM09a} are developed to identify names in emails (target domain) with an underlying statement that two domains have different named-entity distributions. In \cite{nallapati2010blind}, a more difficult situation, i.e., blind domain adaptation, is considered where there is no labeled data observed in the target domain during the model training.

\subsection{Recommender systems}

Cross-domain collaborative filtering (e.g., book recommendation based on music rating information) is an emerging and promising topic in recommender systems \cite{Li:2011:CCF}, because it can improve the performance of recommendation by utilizing observations from other domains. WIth the growth in big data, there is an increasing amount of information from many different sources that is valuable for recommendation, so the need to develop theories and techniques for cross-domain collaborative filtering is urgent. A basic assumption held by  \cite{Iwata2015,7023342} is that latent user-item rating patterns could be used as transferable knowledge between domains, even without any shared users or items \cite{Iwata2015}. The rating matrices from multiple domains are jointly modeled in \cite{ZhangCY10} rather than only focusing on two domains. Except for the rating matrix, there is other important information that could assist recommendation, such as users' implicit feedback (click or not). Transfer Probabilistic Collective Factorization (TPCF) \cite{7023342} unitedly models the rating matrices from two domains with additional auxiliary information. Apart from the shared user-item rating patterns, domain-specific rating patterns are also explicitly expressed in \cite{RenGLG15}.  Rather than recommending items based on rating matrices, CTL in \cite{Tang:2012:CCR} recommends interdisciplinary collaborators to researchers through analyzing research publications from different domains.


\subsection{Text mining}

Text mining, also referred to as text data mining, aims to discover the interest patterns from documents. It comprises text classification, text clustering, sentiment analysis, text retrieval, and so on. Since the successful application of the first two well-known PGMs (i.e., PLSA and LDA) to text mining, there has been a large amount of follow-up work in this area.
One transfer problem in text mining is cross-domain text classification \cite{Xue:2008,WuC10a,Zhuang:2010,Li:2012,ZhuangLYHS13} to classify documents in a (target) domain with a limited number of labeled ones with the help of the other (source) domain with large number of labeled documents. The idea or assumption underpinning this work is that some hidden patterns (e.g., topics) are shared and unchanged between domains.
Another similar transfer problem is cross-collection comparative text mining \cite{Zhai:2004:CMM,5936065,ChenBDXD15} which learns the commonalities and differences between two sets of documents (from two domains). If cross-domain text classification is seen an asymmetric transfer problem, cross-collection comparative text mining can be seen as a symmetric transfer problem.
Short text (target domain) clustering or classification \cite{Jin:2011:TTK,KangML12,5416713} can be seen as a transfer problem because additional related long texts (source domains, e.g., blogs or news) are needed to provide transferable knowledge to assist this problem. The reason why we do not place this work in NLP is because it is a text clustering task rather than simply short text similarity computation.
Sentiment classification is also an interesting problem in text mining, which classifies a document into a sentiment class, such as: positive, negative, or neutral. Cross-domain sentiment classification \cite{GaoL11,WeiHTL2011} is also a transfer problem with the assumption that people tend to use similar words or phrases to express their sentiments even in different domains.
Other examples of transfer problems in text mining are text retrieval \cite{gupta2012bayesian} and topic discovery \cite{liu2014topic}.

\subsection{Computer vision}

Computer vision \cite{forsyth2011computer} is a research area focusing on the modeling, understanding, and utilization of images and videos. PGM has had an enormous impact in this area.
Image categorization is a classical task in image processing, and a corresponding transfer problem of this task is to classify images from different domains, such as object (e.g., \emph{calculator} and \emph{mug}) images from domains: \emph{amazon}, \emph{webcam} and \emph{dslr}. Different domains have different characteristics, i.e., illumination and background \cite{GonenM14}.
One-short/zero-short image categorization \cite{SalakhutdinovTT12,YuA10} is a more complicated transfer problem where the target domain has only one or no image. This problem heavily relies on transferable knowledge from other categories (i.e., domains). The same idea is also applied to object detection \cite{Fei-FeiFP06}.
Image retrieval \cite{gupta2012bayesian} is another classical task in image processing, and its transfer problem is to how to retrieve similar images across different domains (e.g., different animal categories).
For video processing, human pose estimation is a traditional task and has had applied to various scenarios (e.g., video surveillance and games). However, some of the sets of poses have a limited number of labeled data so it is unwise to directly apply traditional models or algorithms to them \cite{LiZZZX13}. Since different poses may share some knowledge, it is more reasonable to utilize this transferable knowledge to estimate the poses with limited training data.
More interestingly, human activity recognition \cite{WeiHTL2011} is facilitated by the transfer from text (source domain) to videos (target domain). This knowledge transfer from domains with completely different natures is quite interesting and promising because it builds a bridge between text and videos.

\subsection{Others}

In addition to the aforementioned four major areas, several other interesting transfer problems have been solved by PGMs in other areas, for example, sequential decision making \cite{WilsonFT12} which increases the interaction between agents through transfer layering; speaker verification \cite{7472720} which transfers knowledge found between a small database (e.g., records of telephone channels) and a large development database; user behavior prediction across multiple social networks (domains) \cite{zhong2012comsoc}, and human categorization \cite{CaniniSG10} which models the transfer behavior of humans on learning different categories. All these diverse transfer problems have shown the capability of PGMs to solve real-world problems and also illustrated of use of transfer PGMs in diverse areas, so it is anticipated that an interesting number of transfer problems which are solved by PGMs will appear in even more diverse areas.

\section{Conclusions and future remarks}
\label{applications}

In this paper, we reviewed the work on probabilistic graphical models for
transfer learning. Since transfer learning can be seen as either a general problem or the methodology by which to resolve this problem, we discussed this work from two aspects: transfer methods and transfer problems. As for transfer methods, existing PGMs for transfer learning have been organized into six categories according to the different characteristics of the shared statistical strengths between the source and target domains, namely Gaussian prior, topic, tree, attribute, factor and model. As for transfer problems, five application areas were identified to cover the existing work, including natural language processing, recommender systems, text mining, computer vision, and others. Through this survey, we found that a number of excellent PGMs have been developed for transfer learning and they have already been successfully applied to resolve various real-world applications. However, compared to the significant success of a large number of PGMs reported in the literature, transfer PGMs are still in their starting phase, and there are quite a few research studies that are worth exploring. Based on the results of this study, several interesting and challenging research points for  further study are as follows: 1) Abstract causal transfer \cite{LuRBY16}. Causal relation learning belongs to Bayesian structure learning, and how to learn the transferable general causal relations from source domain and apply these to the target domain; 2) Deep learning-based transfer learning \cite{Bengio:2011:DLR,7869079}. Deep learning, e.g., deep belief networks \cite{hinton2006fast}, are capable of learning a hierarchial structure containing multiple representations with different granularities. This multi-granularity summarization of data (especially from both domains) could benefit domain adaptation, even with heterogeneous feature spaces; 3) Generative Adversarial Networks-based transfer learning \cite{ChidambaramQ17}. The combination of PGM with discriminative models is also an interesting research direction, where the generalization ability of the PGM is used to control the shared knowledge learning and the prediction ability of discriminative models is used to guide the supervised learning in the target domain.


\vskip 0.2in
\bibliography{CIS}
\bibliographystyle{theapa}

\end{document}